\documentclass[runningheads]{llncs}
\usepackage{graphicx}
\usepackage{booktabs}
\usepackage{array}
\usepackage{bbding} 
\usepackage{tikz}
\usepackage{comment}
\usepackage{amsmath,amssymb} 
\usepackage{color}

\usepackage{colortbl}
\usepackage[
backref,
colorlinks,
linkcolor=red,
anchorcolor=red,
citecolor=green
]{hyperref}

\usepackage[accsupp]{axessibility} 

\usepackage{multirow}
\usepackage{orcidlink}

\pdfoutput=1

\begin{document}
\pagestyle{headings}
\mainmatter
\def\ECCVSubNumber{5661}

\title{UC-OWOD: Unknown-Classified Open World Object Detection} 

\titlerunning{UC-OWOD}
%
\author{Zhiheng Wu\inst{1,2}\orcidlink{0000-0003-2969-6665} \and
Yue Lu\inst{1,2}\orcidlink{0000-0001-7472-9935} \and
Xingyu Chen\inst{3}\orcidlink{0000-0003-3627-0371}
\and
Zhengxing Wu\inst{1,2,\footnotemark[1]}
\orcidlink{0000-0003-2338-5217}
\and \\
Liwen Kang\inst{1,2}\orcidlink{0000-0002-9735-9954}
\and
Junzhi Yu\inst{1,4}\orcidlink{0000-0002-6347-572X}}
\authorrunning{Z. Wu et al.}
%
\institute{Institute of Automation, Chinese Academy of Sciences \\
\and School of Artificial Intelligence, University of Chinese Academy of Sciences \\
\email{\{wuzhiheng2020, luyue2018, zhengxing.wu, kangliwen2020, junzhi.yu\}@ia.ac.cn}
\and Xiaobing.AI \\
\email{chenxingyu@xiaobing.ai} 
\and
Peking University}


\maketitle

\renewcommand{\thefootnote}{\fnsymbol{footnote}}
\footnotetext[1]{Corresponding authors.}

\begin{abstract}
Open World Object Detection (OWOD) is a challenging computer vision problem that requires detecting unknown objects and gradually learning the identified unknown classes. However, it cannot distinguish unknown instances as multiple unknown classes. In this work, we propose a novel OWOD problem called Unknown-Classified Open World Object Detection (UC-OWOD). UC-OWOD aims to detect unknown instances and classify them into different unknown classes. Besides, we formulate the problem and devise a two-stage object detector to solve UC-OWOD. First, unknown label-aware proposal and unknown-discriminative classification head are used to detect known and unknown objects. Then, similarity-based unknown classification and unknown clustering refinement modules are constructed to distinguish multiple unknown classes. Moreover, two novel evaluation protocols are designed to evaluate unknown-class detection. Abundant experiments and visualizations prove the effectiveness of the proposed method. Code is available at \url{https://github.com/JohnWuzh/UC-OWOD}.

\keywords{OWOD, UC-OWOD, Object Detection, Clustering}
\end{abstract}

\section{Introduction}
Nowadays, deep learning methods have achieved great success in object detection~\cite{8237584,DBLP:journals/corr/abs-1708-02002,DBLP:journals/corr/RenHG015,9721820,Cao_2020_CVPR,chen2020joint}. Traditional object detection methods are developed under a closed-world assumption, so they can only detect known (labeled) categories~\cite{Girshick2015FastR,7780460,zhu2020deformable}.
However, the real world contains many unknown (unlabeled) classes that can hardly be properly handled by conventional detection. Therefore, studying the Open World Object Detection (OWOD) problem for detecting unknown instances is of great significance to facilitate the practical application.
 
\begin{figure}[!t]
\centering
\includegraphics[width=1\textwidth]{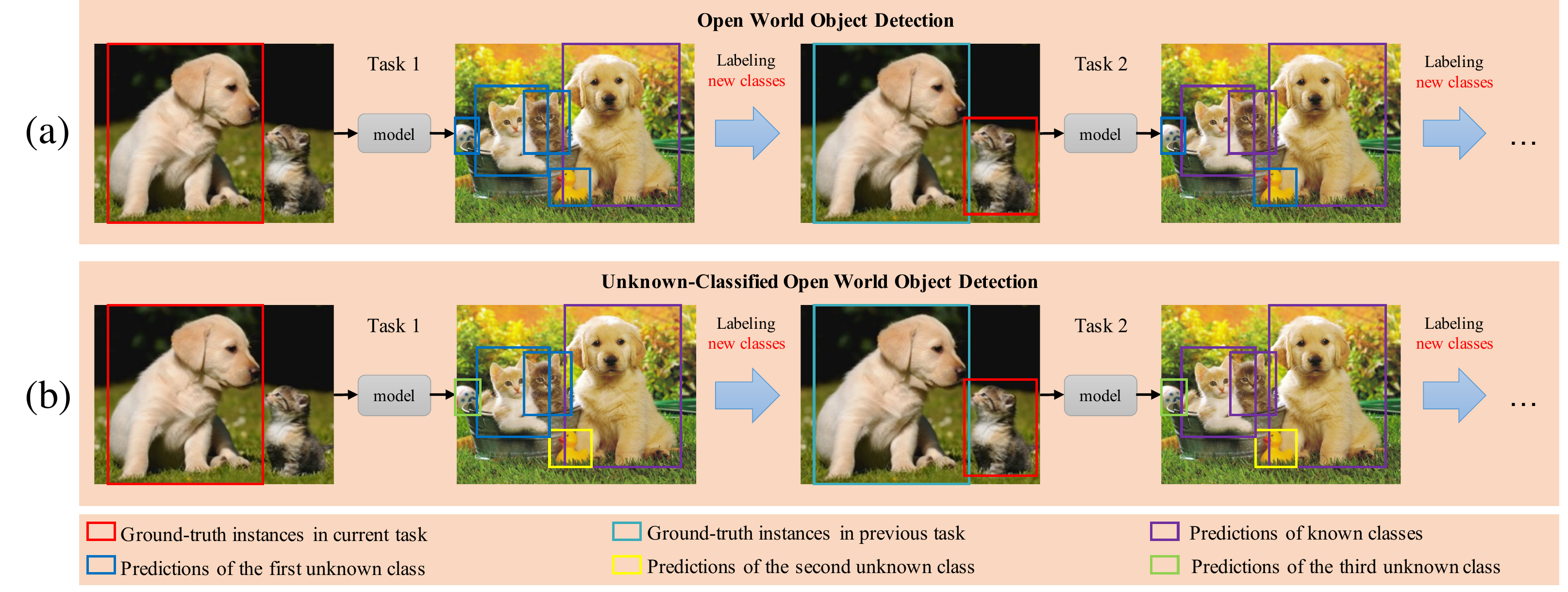}
\caption{The comparison between OWOD and UC-OWOD. They can both learn the newly annotated classes by human annotators without forgetting in the next task. (a) OWOD detects unknown objects as a same class. (b) UC-OWOD can detect unknown objects as different classes.}
\label{fig:intro}
\end{figure}

 The OWOD problem was pioneered by~\cite{joseph2021open}, as shown in Fig.~\ref{fig:intro} (a). OWOD contains multiple incremental tasks. In each task, OWOD is able to identify all unknown instances as \textit{unknown}. Then, human annotators can gradually assign labels to classes of interest, and the model learns these classes incrementally in the next task. However, beyond distinguishing unknown classes, we also need to determine whether multiple unknown instances belong to the same category. Therefore, there is still a huge difficulty when using OWOD for real-world tasks. For example, in practical applications in robotics~\cite{geiger2013vision,lenz2015deep} and self-driving cars~\cite{caesar2020nuscenes,sun2020scalability}, it is necessary to explore the unknown environment and adopt different strategies for different unknown classes, which requires detection algorithms to confidently localize unknown instances and classify them into different unknown classes. 
 
Most existing open-world detectors are designed for OWOD problem. For example, Open World Object Detector (ORE)~\cite{joseph2021open} can detect unknown classes, but it does not consider the case of classifying unknown objects. More specifically, ORE used pseudo-label supervised training to detect unknown instances. Since pseudo-labels can only be marked as \textit{unknown}, the ORE model cannot be directly used to solve the problem of detecting unknown classes as different classes. Similarly, existing OWOD methods models such as \cite{DBLP:journals/corr/abs-2112-01513,zhao2022revisiting} follow ORE's spirit, and we are not aware of any previous work that can distinguish multiple unknown classes. 

Another difficulty in studying unknown object classification problems is the immature evaluation criterion. Existing metrics only evaluate the degree of confusion between unknown and known classes. They cannot evaluate the situation where two unknown objects of different classes are detected as the same class. But these problems cannot be ignored because they may cause the model to misclassify unknown objects. Therefore, a more reasonable evaluation metric is urgently needed to evaluate the detection accuracy of multiple unknown classes.

Considering the above issues, we propose a novel OWOD problem that is closer to the real-world setting, namely, Unknown-Classified Open World Object Detection (UC-OWOD), which can detect unknown objects as different unknown classes (see Fig.~\ref{fig:intro} (b)). Meanwhile, we propose a novel framework based on the two-stage detection pipeline to solve this problem. In particular, we design the unknown label-aware proposal (ULP) to construct unknown object ground-truth, the unknown-discriminative classification head (UCH) to mine unknown objects, the similarity-based unknown classification (SUC) to detect unknown objects as different classes, and the unknown clustering refinement (UCR) to refine the classification of unknown objects. To more accurately evaluate the UC-OWOD problem, we propose novel metrics to evaluate the classification and localization performance of unknown instances. A maximum matching is used to assign ground-truth to unknown objects more reasonably. Ultimately, our model achieves the best performance in both existing evaluation metrics and new evaluation metrics. Our main contributions are as follows:
\begin{itemize}
\item We introduce a new problem setting, i.e., 
unknown-classified open world object detection, to inspire future research on real-world object detection.
\item We propose a method to solve the UC-OWOD problem based on the unknown label-aware proposal, the unknown-discriminative classification head, the similarity-based unknown classification, and the unknown clustering refinement.
\item Novel evaluation metrics for UC-OWOD are proposed, which can evaluate the localization and classification of unknown objects. Extensive experiments are conducted, and the results demonstrate the effectiveness of our method and new metrics for the UC-OWOD problem.
\end{itemize}

\section{Related Work}
\textbf{Open Set Recognition and Detection.} Open Set Recognition was first defined as a constrained minimization problem~\cite{Scheirer2013TowardOS}, and it can submit unknown classes to the algorithm during the testing phase.
It was developed to a multi-class classifier by \cite{10.1007/978-3-319-10578-9_26,scheirer2014probability}. 
Liu et al. considered a long-tailed recognition environment and developed a metric learning framework to identify unseen classes as unknown classes~\cite{DBLP:journals/corr/abs-1904-05160}. 
Self-supervised learning~\cite{Perera_2020_CVPR} and unsupervised learning with reconstruction~\cite{Yoshihashi_2019_CVPR} have also been used for open-set recognition. 
Yue et al. provided a theoretical ground for balancing and improving the seen/unseen classification imbalance~\cite{Yue_2021_CVPR}.
Bendale and Boult proposed a method to adapt deep networks to Open Set Recognition, using OpenMax layers to estimate the probability that the input is from an unknown class~\cite{bendale2016towards}. 
Dhamija et al. first proposed the open-set object detection protocol and formalized the open-set object detection problem~\cite{Dhamija_2020_WACV}. Miller et al. improved object detection performance by extracting label uncertainty under open conditions commonly encountered in robot vision~\cite{DBLP:journals/corr/abs-1710-06677}. Some follow-up work also exploited measures of (spatial and semantic) uncertainty in object detectors to reject unknown categories~\cite{hall2020probabilistic}. Miller et al. found that the correct choice of affinity clustering combinations can greatly improve the effectiveness of classification, spatial uncertainty estimation, and the resulting object detection performance~\cite{miller2019evaluating}. However, these methods cannot gradually adjust their knowledge in a dynamic world. By contrast, our model can dynamically update known classes based on human-annotated labels.\\
\textbf{Open World Recognition and Detection.} Compared to open set problems, open world problem has dynamic datasets and can continuously add new known classes like continuous learning \cite{perez2020incremental,dong2021bridging,rostami2021detection,wang2021wanderlust,kj2021incremental}.
Bendale et al. first proposed Open World Recognition and presented a protocol for the evaluation of open world recognition systems~\cite{DBLP:journals/corr/BendaleB14}.
Xu et al. proposed a meta-learning approach to the open world learning problem that uses only examples of instantly seen classes (including newly added classes) for classification and rejection~\cite{xu2019open}. Joseph et al. presented a new computer vision problem called OWOD~\cite{joseph2021open}. The ORE proposed by them can classify proposals between known and unknown classes, but it relies on a holdout validation set with weak unknown supervision to learn the energy distributions of known and unknown classes. The open-world detection transformer (OW-DETR) improved performance using multi-scale self-attention and deformable receptive fields~\cite{DBLP:journals/corr/abs-2112-01513}. Zhao et al. further proposed an OWOD framework including an auxiliary proposal advisor and a class-specific expelling classifier~\cite{zhao2022revisiting}. None of these methods implements the classification of unknown classes. Our work mainly studies the classification of unknown objects.\\
\textbf{Constrained Clustering.} Constrained clustering is a semi-supervised learning method that involves prior knowledge to assist clustering. The proposed methods for constrained clustering can be divided into three types, i.e., search-based (also known as constraint-based), distance-based (also known as similarity-based), and hybrid (also known as search-and-distance-based) methods~\cite{zhigang2013constrained}. A common technique in search-based methods is to modify the objective function by adding penalty terms for unsatisfied constraints. In distance-based methods, existing clustering methods are usually used, but the distance metric of this method is modified according to prior knowledge. Hybrid methods integrate search-based and distance-based methods. They benefit from the strengths of both and generally perform better than separate methods~\cite{dinler2016survey}. Basu et al. allowed the constraints to be violated with violation cost, while optimizing the distance metric~\cite{199dad7e7703420b9ec7b59548a485fc}. Hsu et al. designed a new loss function to normalize classification with constrained clustering losses, while using other similarity prediction models as pairwise constraints in the clustering process~\cite{hsu2017learning}. Lin et al. took pairwise constraints as prior knowledge to guide the clustering process~\cite{lin2020discovering}. We use pairwise constraints to optimize the unknown object classification in the model.

\section{Unknown-Classified Open World Object Detection}
\subsection{Problem Formulation}
The UC-OWOD problem is defined as follows. There are a set of tasks $\mathcal{T} = \{T_1, T_2,...\}$. In task $T_t$, we have a known class set $\mathcal{K}^{t} = \{1, 2,...,C\}$ and unknown class set $\mathcal{U}^{t} = \{C+1, C+2,...\}$, where $C$ is the number of known classes. The known class set in task $T_{t+1}$ contains that in task $T_t$, i.e., $\mathcal{K}^{t} \subset \mathcal{K}^{t+1}$. The label of the $k$-th object of the known class dataset $\mathcal{D}^{t} = \{ \mathbf{X}^{t},\mathbf{Y}^{t} \} $ is $\mathbf{y}_k = [l_k,x_k,y_k,w_k,h_k], $ where the class label $l_k \in \mathcal{K}^{t}$ and $x_k, y_k, w_k, h_k$ denote the bounding box centre coordinates, width and height, respectively. $\mathbf{X}^{t}$ and $\mathbf{Y}^{t}$ are the input images and labels, respectively. Instances of unknown classes do not have labels. The object detector $\mathcal{M}_C$ is able to identify test instances belonging to any known class, and can also detect the new or unseen class instances as different unknown classes. Human users can identify $u$ new classes of interest from the unknown set of instances $\mathbf{U}^{t}$ and provide the corresponding training examples. Update known class set $\mathcal{K}^{t+1} = \mathcal{K}^{t} \cup \{C+1,...,C+u \}$. By incrementally adding $u$ new classes in the next task, the learner creates an updated model $\mathcal{M}_{C+u}$ without the need to retrain the model on the entire dataset.

\subsection{Overall Architecture}
\begin{figure}[!t]
\centering
\includegraphics[width=1\textwidth]{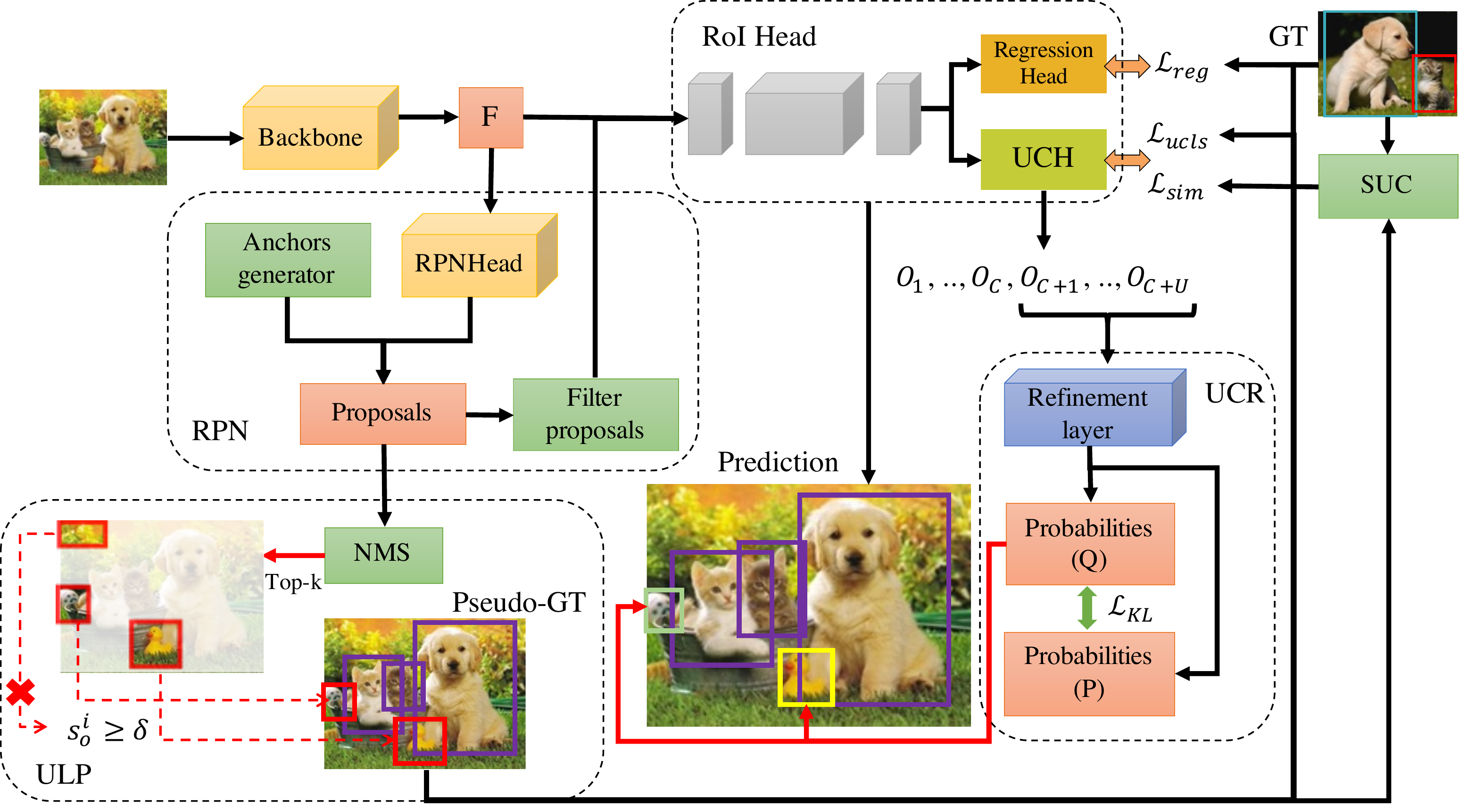}
\caption{The architecture of our model. Pseudo-label candidate boxes are filtered according to whether their score $s_o^i$ is greater than the threshold $\delta$. During the training, ULP constructs Pseudo-GT for unknown objects based on the proposals of RPN. According to the regression head and UCH of the model, $\mathcal{L}_{reg}$ and $\mathcal{L}_{ucls}$ are calculated respectively, and $\mathcal{L}_{sim}$ is got by SUC. During the refining, the unknown objects $\{O_{C+1},...,O_{C+U}\}$ obtained by UCH are input to UCR to refine clustering, where $U$ is the number of unknown classes. }
\label{fig:Architecture}
\end{figure} 
Fig.~\ref{fig:Architecture} shows the overall architecture of the proposed method for UC-OWOD. We use Faster R-CNN~\cite{DBLP:journals/corr/RenHG015} as the base detector. We introduce (1) ULP and UCH to solve the problem of discovering unknown classes from the background, (2) SUC to detect unknown objects as different classes, and (3) UCR to refine the classification of unknown objects and enhance the robustness of the algorithm. In order to model the differences between unknown objects, we propose a new classification loss. Details will be discussed in the following subsections.

\subsection{Detection of Unknown Objects}

\begin{figure}[!t]
\centering
\includegraphics[width=1\textwidth]{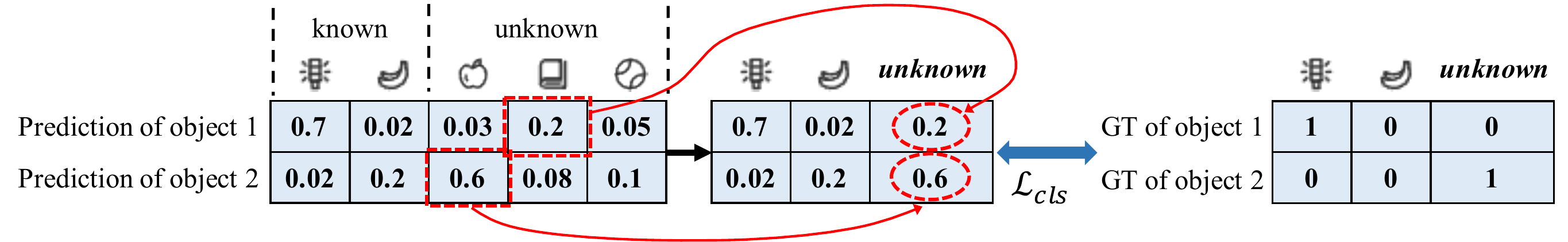}
\caption{The diagram of UCH. Traffic light and banana are known classes. Apple, book, and baseball are unknown classes. The unknown class only selects the value with the highest score when calculating the loss.}
\label{fig:UCH}
\end{figure}

\textbf{Unknown Label-Aware Proposal.} 
Since unknown instances are not labeled, pseudo-labels need to be constructed to train the model's ability to detect unknown classes. We adopt a novel pseudo-labeling strategy, which has better generalization and applicability in detection with multiple unknown classes, as shown in the bottom left of Fig.~\ref{fig:Architecture}. Based on the fact that the Region Proposal Network
(RPN) is class-agnostic, we construct pseudo-labels with bounding box proposals generated by RPN and corresponding objectness scores. First, all proposals are filtered by Non-Maximum Suppression (NMS) to avoid partial overlap between pseudo-labels. Second, we select the filtered top-k background proposals as candidates, which are sorted by their objectness scores. Third, in order to avoid marking the real background regions proposals as \textit{unknown} and make the training results more robust, among the candidates, the proposals with objectness score $s_o$ greater than the threshold $\delta$ are used as pseudo-labels, i.e., $\mathbf{y}_{unk} = [\texttt{unknown}, x_i, y_i, w_i, h_i] $ serves as the unknown label-aware proposal.\\
\textbf{Unknown-Discriminative Classification Head.}
To enable the model to locate and classify unknown classes, we introduce multiple unknown classes in the original classification head: $F_{cls}: \mathbb{R}^D \rightarrow \mathbb{R}^{C+U}$, where $U$ is the number of unknown classes. In the training phase, the pseudo-labels are all marked as $unknown$. The original classification strategy cannot classify a variety of unknown objects, so we modify the original classification loss. As shown in Fig.~\ref{fig:UCH}, the classification loss of unknown classes is computed by using pseudo labels and the maximum probabilities that are predicted by multiple unknown classes. The new classification loss is constructed as
\begin{align}
\mathcal{L}_{ucls} = -\frac{1}{N} \sum_{i=1}^N( \sum_{j=1}^C l_{i,j} \log(p_{i,j}) - l_{i}^* \log( \max \{p_{i,C+1},...,p_{i,C+U}\} ) ),
\end{align}
where $N$ is the number of instances, $l$ is the label of the known class, $l^*$ is the pseudo-label of the unknown class, and $p$ is the predicted probability. 

\begin{figure}[!t]
\centering
\includegraphics[width=1\textwidth]{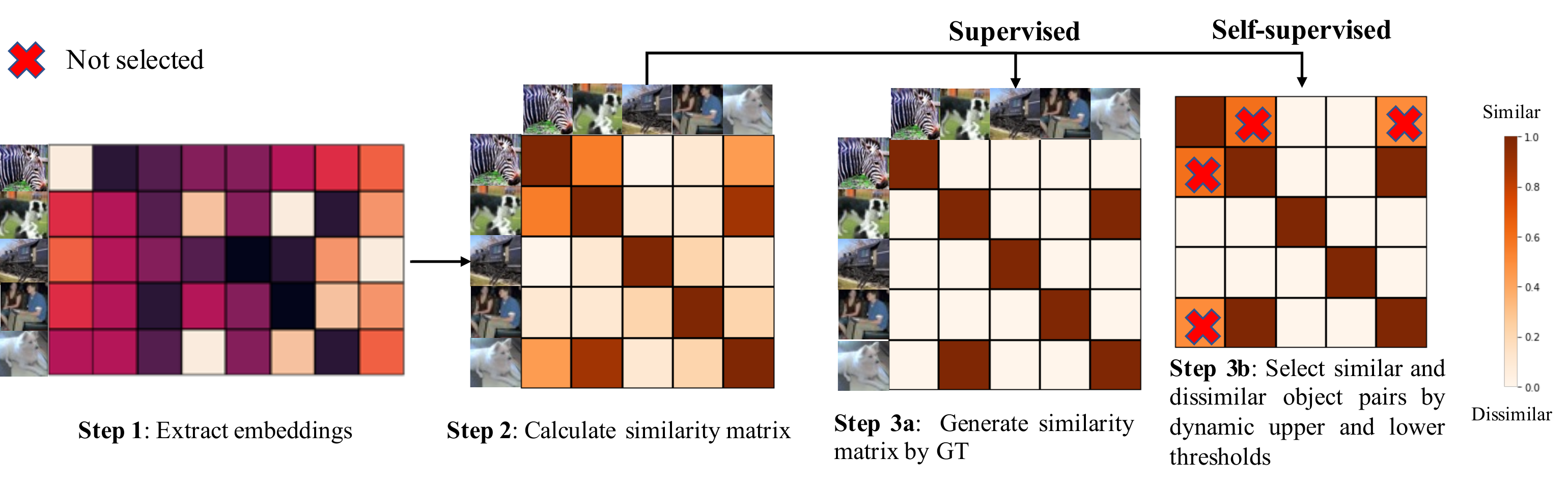}
\caption{Build of similarity matrix of embeddings, using supervised method for known classes and self-supervised method for unknown classes.
}
\label{fig:SUC}
\end{figure}

\subsection{Similarity-Based Unknown Classification}
Clustering unknown classes allow the model to distinguish between different unknown classes. We adopt a pairwise classification loss to measure the similarity between samples. By determining whether pairs of samples are similar, our model can classify unknown classes. The outputs $E$ of the UCH, which can represent category information, are used to compute the similarity matrix $S$:
\begin{align}
 S_{ij} = \frac{E_i E^T_j}{||E_i|| ||E_j||},
\end{align}
where $||\cdot||$ is L2 norm and $i,j \in \{1,\dots,n\}$, and $n$ represents the number of proposals. $ S_{ij} $ represents the similarity between the $i$-th proposal and the $j$-th proposal. As shown in Fig.~\ref{fig:SUC}, we use supervised and self-supervised methods successively to optimize the model.\\
\textbf{Supervised Method.}
We treat labeled data as prior knowledge and use it to guide similar relationships between different unknown instances. In supervised methods, since the relationship between unknown instances is not known, we only use known-known instance pairs, unknown-known instance pairs, known-background instance pairs, and unknown-background instance pairs. We can construct the label matrix $M$ as
\begin{align}
M_{ij} = 
\left\{
\begin{array}{rll}
& 1, \ \text{if}\ l_i = l_j \ \text{and}\ l_i,l_j \notin \mathcal{U}, \\
& 0, \ \text{if}\ l_i \neq l_j, \\
& \text{Not} \ \text{selected}, \ \text{otherwise},
\end{array}
\right.
\end{align}
where $l_i$ is the class label of the $i$-th instance, $i,j \in \{1,\dots,n\}$, and $\mathcal{U}$ is the set of unknown classes. Known instances with ground-truth are utilized to reduce errors.
Therefore, we construct a similarity loss $\mathcal{L}_{sim}$ with labels $M$ and similarity $S$ as
\begin{align}
\mathcal{L}_{sim}(M_{ij},S_{ij}) = -M_{ij}\log(S_{ij})-(1-M_{ij})\log(1-S_{ij}). 
\end{align}
\textbf{Self-Supervised Method.}
We use thresholds to determine whether unknown instance pairs are similar. $TH(\lambda)$ and $TL(\lambda)$ are dynamic upper and lower thresholds applied to the similarity matrix $S$ to obtain the self-labeled matrix $ \tilde{M}$, where $\lambda$ is an adaptive parameter that controls sample selection. Those unknown instance pairs that have similarities between $TH(\lambda)$ and $TL(\lambda)$ are excluded from the training phase. $ \tilde{M}$ is defined as follows:
\begin{align}
\tilde{M}_{ij} = 
\left\{
\begin{array}{rll}
& 1, & \text{if} \ l_i,l_j \in \mathcal{U} \ \text{and} \ S_{ij}>TH(\lambda), \\
& -1, & \text{if} \ l_i,l_j \in \mathcal{U} \ \text{and} \ S_{ij}<TL(\lambda), \\
& 0, & \text{otherwise}.
\end{array}
\right.
\end{align}
Then, we construct the label matrix $\hat{M}$ with the self-labeled matrix $\tilde{M}$ and the class labels $l$ as
\begin{align}
\hat{M}_{ij} = 
\left\{
\begin{array}{rll}
& 1, \ \text{if} \ l_i = l_j\ \text{and} \ l_i,l_j \notin \mathcal{U}, \ \text{or}\ \tilde{M}_{ij}>0,\ \\
& 0, \ \text{if} \ l_i \neq l_j \ \text{or}\ \tilde{M}_{ij}<0, \\
& \text{Not} \ \text{selected}, \ \text{otherwise}.
\end{array}
\right.
\end{align}
The similarity loss $\mathcal{\hat{L}}_{sim}$ is computed by the similarity matrix $S$ and the label matrix $\hat{M}$:
\begin{align}
\mathcal{\hat{L}}_{sim}(\hat{M}_{ij},S_{ij}) = -\hat{M}_{ij}\log(S_{ij})-(1-\hat{M}_{ij})\log(1-S_{ij}) + \mathcal{L}_{ul}(\lambda),
\end{align}
where the penalty term $\mathcal{L}_{ul}(\lambda)$ for the number of samples is given as
\begin{align}
\mathcal{L}_{ul}(\lambda) = TH(\lambda)-TL(\lambda).
\end{align}
The adaptive parameter $\lambda$ updated by:
\begin{align}
\lambda:=\lambda-\eta \cdot \frac{\partial \mathcal{L}_{ul}(\lambda)}{\partial \lambda},
\end{align}
where $\eta$ is the learning rate of $\lambda$. 
More and more instance pairs participate in the training phase as $\lambda$ is updating. To obtain clustering-friendly representations, we train the model from easily classified unknown instance pairs to hardly classified unknown instance pairs iteratively as the thresholds change.
The iterative process is terminated when $TH(\lambda) \leq TL(\lambda) $.
\subsection{Unknown Clustering Refinement}
To enhance the robustness of the proposed algorithm, we apply the soft assignment method~\cite{xie2016unsupervised} to improve the unknown classification based on the previous network output. UCR uses clustering to improve the separability of unknown objects. In the first step, according to the output of UCH, the embedding $E$ of the unknown class and the cluster centroid $\Phi$ of the unknown class are obtained. And we compute a soft assignment between $E_i$ and $\Phi_j$ saved in the refinement layer while using the Student’s t-distribution~\cite{van2008visualizing} as the kernel: 
\begin{align}
P_{ij} = \frac{(1+||E_i-\Phi_j||^2)^{-1}}{\sum_{k}(1+||E_i-\Phi_k||^2)^{-1}},
\end{align}
where $P_{ij}$ can be interpreted as the probability (soft assignment) of assigning instance $i$ to cluster $j$. In the second step, the auxiliary target distribution $Q$ is used to refine the clusters based on their high confidence assignments:

\begin{align}
Q_{ij} = \frac{P_{ij}^2 / F_i}{\sum _{k}P_{ik}^2 / F_{k}},
\end{align}
where $ F_i = \sum _i P_{ij} $ is soft cluster frequencies. The quadratic term of the auxiliary target distribution can emphasize high confidence assignments. Therefore, with the assistance of the auxiliary target distribution, the model can gradually learn good clustering structure and improve clustering purity.
Then, we minimize the Kullback-Leibler (KL) divergence loss between the soft assignments $P$ and the auxiliary distribution $Q$ to refine clustering:

\begin{align}
\mathcal{L}_{KL}=\operatorname{KL}(Q||P) = \sum_i \sum_j Q_{ij} \log \frac{Q_{ij}}{P_{ij}}.
\end{align}

\subsection{Training and Refining}
\textbf{Training.} Our model is trained end-to-end with the following loss function:
\begin{align}
\mathcal{L}_{tra} =\alpha_{1} \mathcal{L}_{rpn} + \alpha_{2} \mathcal{L}_{ucls} + \alpha_{3} \mathcal{L}_{reg} + \alpha_{4} \mathcal{L}_{sim},
\end{align}
where $ \mathcal{L}_{rpn} $ and $\mathcal{L}_{reg}$ denote the loss terms for RPN and bounding box regression, respectively. In detail, $ \mathcal{L}_{rpn} $ is formulated
using the standard RPN loss~\cite{DBLP:journals/corr/RenHG015}, $\mathcal{L}_{reg}$ is the
standard $\ell_{1}$ regression loss. $\alpha_{1}, \alpha_{2}, \alpha_{3}, \alpha_{4}$ denote weight factors. 
When the model is only trained with the current class of task $T_t$, it will catastrophically forget the information learned in the previous task~\cite{ffe0793b43f842d2a50467d736a80c83,French1999CatastrophicFI}. Comparing existing solutions, i.e. parameter regularization~\cite{DBLP:journals/corr/abs-1711-09601,8107520}, exemplar replay~\cite{DBLP:journals/corr/RebuffiKL16,DBLP:journals/corr/abs-1807-09536}, dynamically expanding networks~\cite{DBLP:journals/corr/abs-1711-05769,DBLP:journals/corr/RusuRDSKKPH16,DBLP:journals/corr/abs-1801-01423}, and meta-learning~\cite{Rajasegaran_2020_CVPR,NEURIPS2020_a5585a4d}, we choose a relatively simple few-example replay method~\cite{DBLP:journals/corr/abs-2003-06957,prabhu2020greedy,joseph2021open}. The model is finetuned using a set of stored examples for each known class after learning the task $T_t$. 
\\
\textbf{Refining.} In the clustering refinement stage for unknown objects, the main purpose is to improve the classification of unknown objects. We only use the KL divergence loss for training on unknown objects:

\begin{align}
\mathcal{L}_{ref} = \mathcal{L}_{KL}.
\end{align}

\section{Experiments}

\subsection{Preparation} 
\textbf{Datasets.} We evaluate our model for the UC-OWOD problem on the set of tasks $\mathcal{T}=\{T_1,T_2,\cdots \}$. Classes in $T_\lambda$ are introduced when $t=\lambda$.
For the task $T_t$, all introduced classes in $\{ T_{\tau} : \tau \leq t\}$ are $known$ and classes in $\{ T_{\tau} : \tau > t\}$ are $unknown$.
As shown in Table~\ref{table:datasets}, we construct 4 tasks with 20 classes in each task using the Pascal VOC~\cite{88a29de36220442bab2d284210cf72d6} and MS-COCO~\cite{10.1007/978-3-319-10602-1_48} datasets.
The task $T_1$ consists of all VOC classes and data, which do not contain any information about unknown instances. This allows the model to be tested without any \textit{unknown} information during the training phase. 
The remaining 60 classes of MS-COCO are divided into three parts, i.e., $T_2$, $T_3$, and $T_4$.
Although the training images in $T_2$ and $T_3$ do not have labels of unknown instances, they contain unknown instances, which can test the effect of the model in this situation. In every task, the evaluation data consists of Pascal VOC test split and MS-COCO validation split.
\setlength{\tabcolsep}{4pt}
\begin{table}[!t]
\begin{center}
\caption{Datasets for each task. Table shows the semantics and the number of images and instances each task contains.}
\label{table:datasets}
\begin{tabular}{c|cccc}
\toprule[1pt]
Task& Task 1 & Task 2 & Task 3& Task 4 \\
\midrule[0.5pt]
\multicolumn{1}{c|}{\multirow{2}{*}{Semantic split}} & VOC & Outdoor, Accessories & Sports & Electronic, Indoor \\
\multicolumn{1}{c|}{}&Classes & Appliance, Truck & Food& Kitchen, Furniture \\
\midrule[0.5pt]

Training images & 16551& 45520& 39402 & 40260\\
\midrule[0.5pt]
Training instances& 47223& 113741 & 114452& 138996 \\
\midrule[0.5pt]
Test images & \multicolumn{4}{c}{10246}\\
\midrule[0.5pt]
Test instances& \multicolumn{4}{c}{61707} \\
\bottomrule[1pt]
\end{tabular}
\end{center}
\end{table}
\setlength{\tabcolsep}{1.4pt}\\
\textbf{Evaluation Metrics.} 
For the overall evaluation of unknown classes, we use two evaluation metrics, i.e., Absolute Open-Set Error (A-OSE)~\cite{DBLP:journals/corr/abs-1710-06677,joseph2021open} and Wilderness Impact (WI)~\cite{Dhamija_2020_WACV,joseph2021open}. 
A-OSE is the number of unknown objects misclassified into \textit{known}. WI is calculated by true positive proposals $TP_{\mathcal{K}}$ and false positive proposals $FP_{\mathcal{K}}$ of current \textit{known}:
\begin{align}
\operatorname{WI} = \frac{\operatorname{A-OSE}}{TP_{\mathcal{K}} + FP_{\mathcal{K}}}.
\end{align}
For the refinement of unknown classes, there is no label-prediction pair, so mean average precision (mAP) does not work. We are also not aware of any other metric that can handle the evaluate multiple unknown categories. Inspired by the clustering evaluation metric, i.e., clustering accuracy~\cite{DBLP:journals/corr/abs-2109-01306}, we introduce a novel evaluation metric, unknown mean average precision (UC-mAP), to evaluate the detection of unknown classes. Therefore, UC-mAP is the mAP with automatic category matching:
\begin{align}
\operatorname{UC-mAP}(\mathcal{Y}_{gt},\mathcal{Y}_{pre}) =\max \limits_{perm \in \mathcal{P}} \operatorname{mAP}(perm(\mathcal{Y}_{pre}),\mathcal{Y}_{gt}),
\end{align}
where $\mathcal{P}$ is the set of all permutations in of $1$ to $U$, $U$ is the number of unknown classes, $\mathcal{Y}_{pre}$ is the predicted value, and $\mathcal{Y}_{gt}$ is the ground-truth. The best match uses the Hungarian algorithm \cite{kuhn1955hungarian} for fast computation. The model is also better if it can detect some new instances which are unlabeled in the MS-COCO dataset, but traditional mAP metrics are very sensitive to missing annotations and treat such detections as false positives. Therefore, we also use the unknown class Recall~\cite{DBLP:journals/corr/LuKBL16,DBLP:journals/corr/abs-1804-04340,DBLP:journals/corr/abs-2112-01513} after maximum matching as the evaluation metric, i.e., UC-Recall.
\begin{figure}[!t]
\centering
\includegraphics[width=1\textwidth]{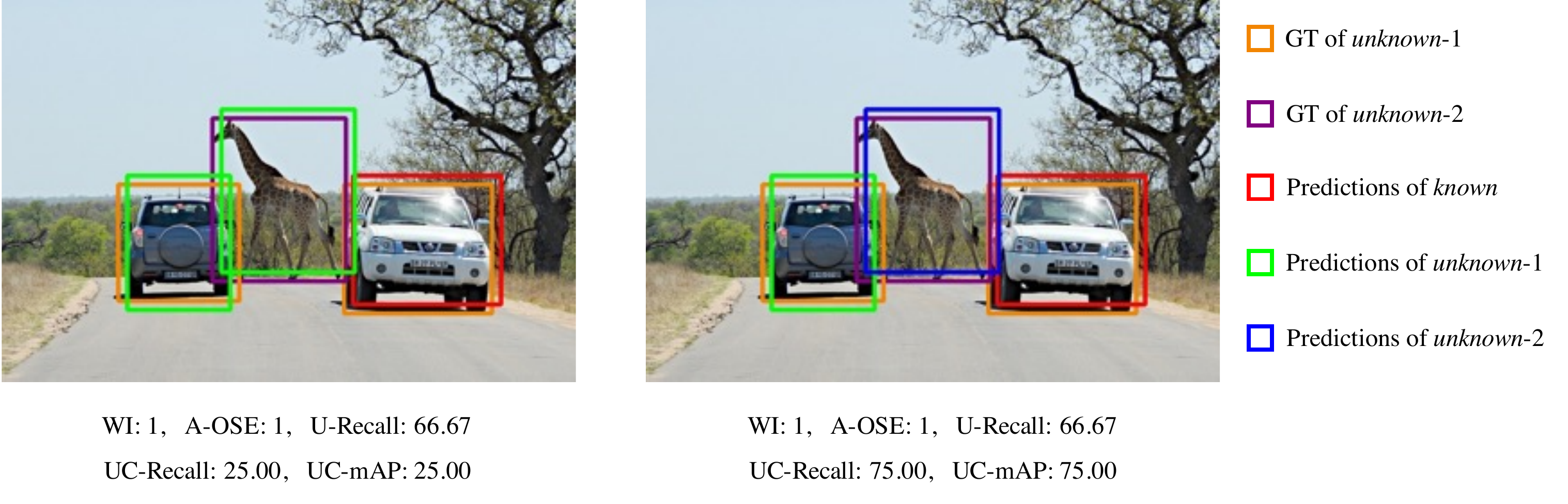}
\caption{\textit{Car} is \textit{unknown}-1 and \textit{giraffe} is \textit{unknown}-2. Both images mis-detect the car on the right as \textit{known}, and the left image mis-detects the giraffe as \textit{unknown}-1.}
\label{fig:umap}
\end{figure}
\setlength{\tabcolsep}{4pt}
\begin{table}[!t]
\begin{center}
\caption{Validation results of UC-mAP and UC-Recall. Label-free UC-mAP can achieve the same evaluation results as label-based mAP, as does Recall.}
\label{table:ucmap}
\begin{tabular}{c|c|c|c|c}
\toprule[1pt]

 &Task 1&Task 2&Task 3&Task 4\\

\midrule[0.5pt]
 mAP / Recall& 56.43 / 75.40 & 46.14 / 68.02& 28.03 / 53.26& 26.72 / 50.88 \\
 UC-mAP / UC-Recall& 56.43 / 75.40 & 46.14 / 68.02& 28.03 / 53.26& 26.72 / 50.88 \\
\bottomrule[1pt]

\end{tabular}
\end{center}
\end{table}
\setlength{\tabcolsep}{1.4pt}

\noindent \textbf{Implementation Details.} 
Our model is based on the standard Faster R-CNN~\cite{DBLP:journals/corr/RenHG015} object detector with ResNet-50~\cite{7780459} backbone. We set the total number of unknown and known classes to 80, which corresponds to the MS-COCO dataset. As described earlier, in the classification loss, we only learn the unknown class with the highest prediction probability. This is achieved by setting the logits of the invisible classes to a large negative value ($v$) so that their contribution to the softmax is negligible ($e^{-v} \rightarrow 0 $). We set $TH(\lambda) = 0.95 - \lambda,~TL(\lambda) = 0.455+0.1\lambda,~\alpha_{1} = \alpha_{2} = \alpha_{3} = 1, \alpha_{4} = 0.5$, and learning rate is 0.01. When refining, we fix the layers before the refinement layer and use a learning rate of 0.1. The initial cluster centroids of unknown classes are obtained using K-means. Because the refinement phase relies on the unknown object information in the training set, we only use UCR for task 2 and task 3.\\
\textbf{Validity of UC-mAP and UC-Recall.} 
We analyze the evaluation results of WI, A-OSE, U-Recall, UC-Recall and UC-mAP in different situations (see Fig.~\ref{fig:umap}). All metrics reflect the situation where unknown objects are misclassified as known. WI, A-OSE, and U-Recall~\cite{DBLP:journals/corr/abs-2112-01513} cannot determine whether \textit{unknown}-1 and \textit{unknown}-2 are wrongly classified into the same class, but UC-Recall and UC-mAP may result in higher scores under correct detection.
UC-Recall and UC-mAP are further evaluated with known classes of \textit{Oracle} detectors, which can access to all known and unknown labels at any task (see Table~\ref{table:ucmap}).
We can see that UC-mAP/UC-Recall are equivalent to mAP/Recall when the model is trained with the corresponding labels. 
\setlength{\tabcolsep}{4pt}
\begin{table}[!t]
\begin{center}
\caption{The performance of our model on known classes. $PK$ means the mAP of previously known instances and $CK$ means the mAP of current known instances.}
\label{table:known}
\begin{tabular}{c|c|c|c|c}
\toprule[1pt]

& Task 1& Task 2& Task 3& Task 4\\

\midrule[0.5pt]

$PK$ ($\uparrow$) / $CK$ ($\uparrow$)& - / 50.66& 33.13 / 30.54& 28.80 / 16.34& 25.57 / 15.88\\ 
\bottomrule[1pt]
\end{tabular}
\end{center}
\end{table}
\setlength{\tabcolsep}{1.4pt}
\setlength{\tabcolsep}{4pt}
\begin{table}[!t]
\begin{center}
\caption{The performance of our model on UC-OWOD. WI, A-OSE, UC-mAP and UC-Recall quantify how the model handles unknown classes.}
\label{table:ucowod}
\begin{tabular}{ccccccccc}
\toprule[1pt]
Task 1 & & Oracle & Faster-RCNN & \begin{tabular}[c]{@{}c@{}}Faster-RCNN\\ +Finetuning\end{tabular} & ORE & Ours & Ours+UCR \\ 
\midrule[0.5pt]
 UC-mAP& ($\uparrow$) & 0 & 0 & - & 0.0133 & \textbf{0.1344} &- \\ 
WI& ($\downarrow$) & - & 0.0188&-& 0.0155& \textbf{0.0136} & - \\ 
A-OSE& ($\downarrow$) & - & 13300 & - & 10672 & \textbf{9294} & - \\ 
UC-Recall& ($\uparrow$) &- & 0 & - & 0.7772 & \textbf{2.3915} & - \\ 
\bottomrule[1pt]
 \noalign{\smallskip}
\toprule[1pt]
Task 2 & & Oracle & Faster-RCNN & \begin{tabular}[c]{@{}c@{}}Faster-RCNN\\ +Finetuning\end{tabular} & ORE & Ours & Ours+UCR \\ 
\midrule[0.5pt]
 UC-mAP& ($\uparrow$) & 15.50 & 0 & 0 & 0.0065 & 0.0862 &\textbf{0.1694} \\ 
WI& ($\downarrow$) & 0.0022 & 0.0069 & 0.0140 &0.0153 & 0.0116 & 0.0117 \\ 
A-OSE& ($\downarrow$) & 6050 & 4582 & 7169 &10376 & 5602 & 5602 \\ 
UC-Recall& ($\uparrow$) & 40.45 & 0 & 0 & 0.0371 & 2.6926 & \textbf{3.4431} \\ 
\bottomrule[1pt]
 \noalign{\smallskip}
\toprule[1pt]
Task 3 & & Oracle & Faster-RCNN & \begin{tabular}[c]{@{}c@{}}Faster-RCNN\\ +Finetuning\end{tabular} & ORE & Ours & Ours+UCR \\ 
\midrule[0.5pt]
 UC-mAP &($\uparrow$)& 10.61 & 0 & 0 & 0.0070 & 0.0249 & \textbf{0.0744} \\ 
WI& ($\downarrow$) & 0.0042 & 0.0241 & 0.0099 & 0.0086 & 0.0073 & \textbf{0.0073} \\ 
A-OSE& ($\downarrow$) & 4857 & 4841 & 9181 & 7544 & 3801 & \textbf{3801} \\ 
UC-Recall& ($\uparrow$) &28.54 & 0 & 0 & 0.8833 & 4.8077 & \textbf{8.7303} \\ 
\bottomrule[1pt]
\end{tabular}

\end{center}
\end{table}
\setlength{\tabcolsep}{1.4pt}

\begin{figure}[!t]
\centering
\includegraphics[width=1\textwidth]{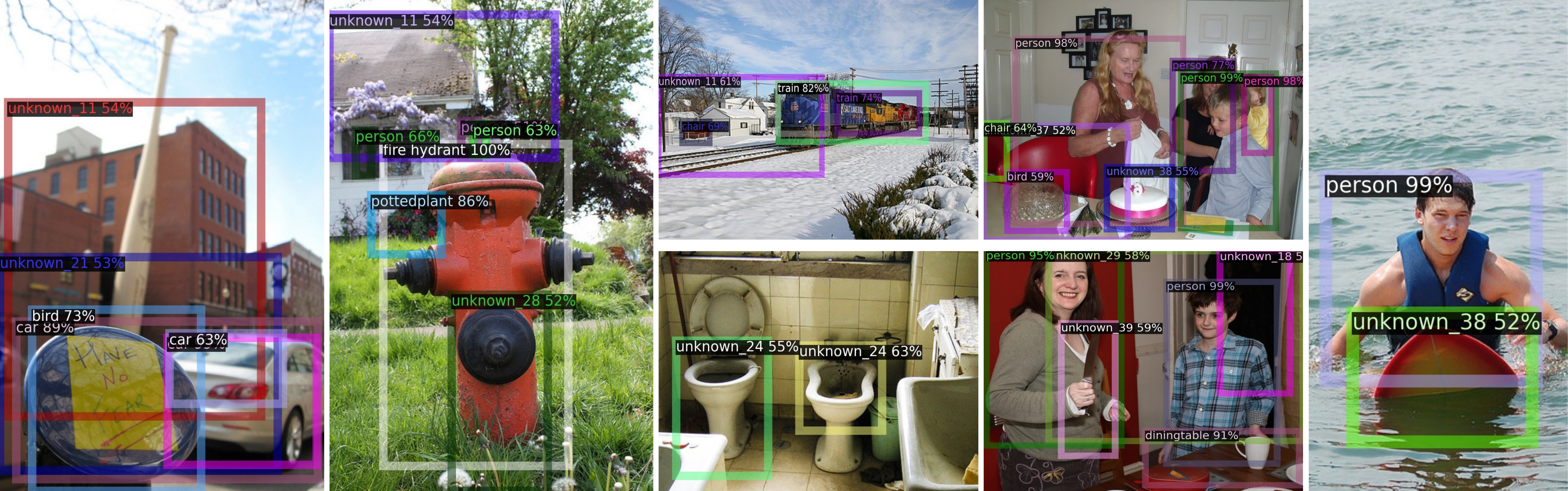}
\caption{Qualitative results of our model. \textit{unknown}\_$x$ represents an unknown object of the $x$-th class. Our model detects the \textit{house} as \textit{unknown}-11 and is able to distinguish it from other unknown classes in the same image. This means that our model cannot only detect the categories annotated in the MS-COCO dataset, but also mine new categories and distinguish them from other categories. Some other unknown classes are also shown, i.e., $toilet$ as \textit{unknown}-24, \textit{knife} as \textit{unknown}-39 and so on. The last column shows a failure case that misclassify \textit{surfboard} as \textit{unknown}-38 which was actually \textit{cake}. The Supplementary Materials contain more visualised results.}
\label{fig:result}
\end{figure}
\subsection{Results and Analysis}\label{subsection:results}
As shown in Table~\ref{table:known}, our model is able to avoid catastrophic forgetting of previous classes. 
To better analyze the performance on the UC-OWOD problem, we compare our model with Faster-RCNN and ORE, whose performance on unknown object detection is shown in Table~\ref{table:ucowod}. Due to limited space, full experimental data are given in Supplementary Materials. WI and A-OSE metrics are used to quantify the degree of confusion between unknown instances and any known classes. The UC-Recall metric is used to quantify the ability of the model to retrieve unknown object instances. The UC-mAP metric is used to quantify the average level of detection of all unknown classes by the model. Under the setting of UC-OWOD, Faster-RCNN and Faster-RCNN
+Finetuning do not have the ability to detect unknown instances, and finetuning will result in lower scores for WI and A-OSE. In all tasks, we achieve better results than ORE on measures about unknown classes. The ability of the model to detect unknown objects is significantly improved after adding UCR. Fig.~\ref{fig:result} and Supplementary Materials show qualitative results on example images. 
\setlength{\tabcolsep}{4pt}
\begin{table}[!t]
\begin{center}

\caption{Ablation experimental results of our model.}
\label{table:ablation}
\begin{tabular}{c|cc|cccc}
\toprule[1pt]

ID& UCH & SUC & WI ($\downarrow$) &A-OSE ($\downarrow$)& UC-Recall ($\uparrow$)& UC-mAP ($\uparrow$) \\

\midrule[0.5pt]

1& \XSolidBrush & \XSolidBrush &0.0213 & 16453& 0& 0\\
2& \XSolidBrush &\CheckmarkBold &0.0247 & 16667& 0 & 0 \\
3&\CheckmarkBold & \XSolidBrush & 0.0155&\textbf{9185} & 1.4796 & 0.0176 \\
4& \CheckmarkBold &\CheckmarkBold & \textbf{0.0136} & 9294 &\textbf{2.3915} &\textbf{0.1343}\\

\bottomrule[1pt]
\end{tabular}
\end{center}
\end{table}
\setlength{\tabcolsep}{1.4pt}
\setlength{\tabcolsep}{4pt}
\begin{table}[!t]
\begin{center}
\caption{Sensitivity analysis on hyperparameters.}
\label{table:parameters}
\begin{tabular}{c|ccc|ccccc}
\toprule[1pt]

\multirow{2}{*}{ID}& NMS & \multirow{2}{*}{$\delta$} &Number of &WI &A-OSE& UC-Recall & UC-mAP \\
&threshold& & pseudo-GT & ($\downarrow$) & ($\downarrow$) & ($\uparrow$) & ($\uparrow$) \\

\midrule[0.5pt]
1& 0.3 & 0.3 & 1& 0.0146 & 12649 & 0.9314 & 0.0375\\
2& 0.3& 0.3 & 5 & 0.0136 & \textbf{9294} & \textbf{2.3915} & \textbf{0.1343} \\
3& 0.3 & 0.7 & 1& \textbf{0.0101} & 11323 & 1.9017 & 0.0995\\
4& 0.3 & 0.7 & 5 & 
0.0143&
10161&
1.6730&
0.1212
\\
5& 0.7 & 0.3 & 1 &
0.0203&
12243&
0.7222&
0.0037
\\
6& 0.7 & 0.3 & 5&
0.0202&
12780&
0.7923&
0.0120
\\
7& 0.7 & 0.7 & 1&
0.0240&
13032&
0.1399&
0.0004
\\
8& 0.7 & 0.7 & 5&
 0.0141&
12156&
0.8827&
0.0026
\\
\bottomrule[1pt]
 
\end{tabular}
\end{center}
\end{table}
\setlength{\tabcolsep}{1.4pt}
\subsection{Ablation Study} 
\textbf{Ablation of Components.} 
We design ablation experiments to study the contributions of UCH and SUC in the model (see Table~\ref{table:ablation}).
When UCH and SUC (row 1 and row 2) are missing, the model loses its ability to detect unknown classes. Adding only SUC (row 2) will not improve the model's ability to detect unknown classes. 
Only the absence of SUC (row 3) affects the classification ability for unknown classes, but the model performs best at the detection of known classes. 
Hence, the scores of WI, UC-Recall, and UC-mAP are worse than those that have both UCH and SUC (row 4). Therefore, the best performance is achieved when both components are present.\\
\textbf{Sensitivity Analysis on Hyperparameters.} 
As shown in Table~\ref{table:parameters}, we analyze the detection performance of the model under different hyperparameter settings.
When the NMS threshold is large, the recall rate for unknown classes is low, because the model may set the region with a high degree of coincidence with the known class label as a pseudo label. The model can only locate known instance regions, but cannot locate unknown instance regions. When the value of $\delta$ is large, the model tends to label fewer unknown classes, resulting in poorer detection performance of the model for unknown classes. 
Similarly, when the Number~of~pseudo-GT is set to 1, the model will be less effective due to fewer unknown classes being labeled. 
We chose the hyperparameter settings with the better scores for WI, A-OSE, UC-Recall, and UC-mAP, i.e., NMS~threshold is 0.3, $\delta$ is 0.3, and Number~of~pseudo-GT is 5.

\section{Conclusions and Future Work}
In this work, we have proposed a novel problem UC-OWOD on the basis of OWOD, which is closer to the real world. The UC-OWOD requires detecting unknown objects as different unknown classes. We also establish evaluation protocols for this issue. In addition, we propose a new method including ULP, UCH, SUC, and UCR. Abundant experiments demonstrate the effectiveness of our method on the UC-OWOD problem and also verify the rationality of our metrics. In future work, we hope to apply our method to some real-world online tasks and achieve open-world automatic annotation.\\
\noindent \textbf{Acknowledgements.} This work was supported in part by the National Key Research and Development Program of China under Grant 2019YFB1310300 and in part by the National Natural Science Foundation of China under Grant 62022090.

\clearpage
%
%


\pagestyle{headings}
\mainmatter
\def\ECCVSubNumber{5661} 

\title{UC-OWOD: Unknown-Classified Open World Object Detection (Supplementary Materials)} 

\titlerunning{UC-OWOD}
%
\author{Zhiheng Wu\inst{1,2}\orcidlink{0000-0003-2969-6665} \and
Yue Lu\inst{1,2}\orcidlink{0000-0001-7472-9935} \and
Xingyu Chen\inst{3}\orcidlink{0000-0003-3627-0371}
\and
Zhengxing Wu\inst{1,2,\footnotemark[1]}
\orcidlink{0000-0003-2338-5217}
\and \\
Liwen Kang\inst{1,2}\orcidlink{0000-0002-9735-9954}
\and
Junzhi Yu\inst{1,4}\orcidlink{0000-0002-6347-572X}}
\authorrunning{Z. Wu et al.}
%
\institute{
Institute of Automation, Chinese Academy of Sciences \\
\and
School of Artificial Intelligence, University of Chinese Academy of Sciences\\
\and Xiaobing.AI \\
\and
Peking University
}
\maketitle

\setlength{\tabcolsep}{4pt}
\begin{table}[!htp]
\begin{center}

\caption{The comparison of \textit{Oracle}, ORE, and our model on UC-OWOD. WI, A-OSE, UC-mAP and UC-Recall reflect how the model handles unknown classes, and mAP measures the ability to detect known classes. It can be seen that our model far outperforms other models in handling unknown classes.}
\label{table:qr}
\resizebox{\textwidth}{!}{
\begin{tabular}{ccccccccc}

\toprule[1pt]
 \noalign{\smallskip}
Task 1 & & Oracle & Faster-RCNN & \begin{tabular}[c]{@{}c@{}}Faster-RCNN\\ +Finetuning\end{tabular} & ORE & Ours & Ours+UCR \\ 
\noalign{\smallskip}
\midrule[0.5pt]
\noalign{\smallskip}
mAP ($\uparrow$)& \begin{tabular}[c]{@{}c@{}}Current\\ known\end{tabular} & 56.49 & 55.38 & - & 56.34 & 50.66 & -\\ 
 \noalign{\smallskip}

 UC-mAP& ($\uparrow$) & 0 & 0 & - & 0.0133 & \textbf{0.1344} &- \\

 \noalign{\smallskip}
WI& ($\downarrow$) & - & 0.0188 & - & 0.0155 & \textbf{0.0136} & - \\ 
 \noalign{\smallskip}

A-OSE& ($\downarrow$) & - & 13300 & - & 10672 & \textbf{9294} & - \\ 
 \noalign{\smallskip}

UC-Recall& ($\uparrow$) &- & 0 & - & 0.7772 & \textbf{2.3915} & - \\ 
 \noalign{\smallskip}
\bottomrule[1pt]
 \noalign{\smallskip}

\toprule[1pt]

 \noalign{\smallskip}
Task 2 & & Oracle & Faster-RCNN & \begin{tabular}[c]{@{}c@{}}Faster-RCNN\\ +Finetuning\end{tabular} & ORE & Ours & Ours+UCR \\ 
\noalign{\smallskip}
\midrule[0.5pt]
\noalign{\smallskip}
\multirow{5}{*}{mAP ($\uparrow$)}& \begin{tabular}[c]{@{}c@{}}Previously\\ known\end{tabular} & 54.83 & 0 & 40.90 & 52.27 & 33.13 & 33.13 \\ 
 \noalign{\smallskip}

& \begin{tabular}[c]{@{}c@{}}Current\\ known\end{tabular} & 37.92 & 36.15 & 31.60 & 25.49 & 30.54 & 30.54\\ 
 \noalign{\smallskip}

 & Both & 46.37 & 18.07 & 36.25 &38.88 & 31.84 &31.84 \\ 
 \noalign{\smallskip}

 UC-mAP& ($\uparrow$) & 15.50 & 0 & 0 & 0.0065 & 0.0862 &\textbf{0.1694} \\ 
 \noalign{\smallskip}

WI& ($\downarrow$) & 0.0022 & 0.0069 & 0.0140 &0.0153 & 0.0116 & 0.0117 \\ 
 \noalign{\smallskip}

A-OSE& ($\downarrow$) & 6050 & 4582 & 7169 &10376 & 5602 & 5602 \\ 
 \noalign{\smallskip}

UC-Recall& ($\uparrow$) & 40.45 & 0 & 0 & 0.0371 & 2.6926 & \textbf{3.4431} \\ 
 \noalign{\smallskip}
\bottomrule[1pt]
 \noalign{\smallskip}

\toprule[1pt]
 \noalign{\smallskip}
Task 3 & & Oracle & Faster-RCNN & \begin{tabular}[c]{@{}c@{}}Faster-RCNN\\ +Finetuning\end{tabular} & ORE & Ours & Ours+UCR \\ 
\noalign{\smallskip}
\midrule[0.5pt]
\noalign{\smallskip}
\multirow{5}{*}{mAP ($\uparrow$)}& \begin{tabular}[c]{@{}c@{}}Previously\\ known\end{tabular} & 30.77 & 0 & 30.55 & 38.45 & 28.80 & 28.80 \\ 
 \noalign{\smallskip}

& \begin{tabular}[c]{@{}c@{}}Current\\ known\end{tabular} & 22.56 & 19.78 & 18.16 & 12.65 & 16.34 & 16.34\\ 
 \noalign{\smallskip}

 & Both & 28.03 & 6.59 & 26.42 & 29.85 & 24.65 & 24.65 \\ 
 \noalign{\smallskip}

 UC-mAP &($\uparrow$)& 10.61 & 0 & 0 & 0.0070 & 0.0249 & \textbf{0.0744} \\ 
 \noalign{\smallskip}

WI& ($\downarrow$) & 0.0042 & 0.0241 & 0.0099 & 0.0086 & 0.0073 & \textbf{0.0073} \\ 
 \noalign{\smallskip}

A-OSE& ($\downarrow$) & 4857 & 4841 & 9181 & 7544 & 3801 & \textbf{3801} \\ 
 \noalign{\smallskip}

UC-Recall& ($\uparrow$) &28.54 & 0 & 0 & 0.8833 & 4.8077 & \textbf{8.7303} \\ 
 \noalign{\smallskip}
\bottomrule[1pt]
 \noalign{\smallskip}

\toprule[1pt]
 \noalign{\smallskip}
 Task 4 & & Oracle & Faster-RCNN & \begin{tabular}[c]{@{}c@{}}Faster-RCNN\\ +Finetuning\end{tabular} & ORE & Ours & Ours+UCR \\ 
\noalign{\smallskip}
\midrule[0.5pt]
\noalign{\smallskip}
\multirow{5}{*}{mAP ($\uparrow$)}& \begin{tabular}[c]{@{}c@{}}Previously\\ known\end{tabular} & 29.18 & 0 & 24.74 & 30.08 & 25.57 & -\\ 
 \noalign{\smallskip}

& \begin{tabular}[c]{@{}c@{}}Current\\ known\end{tabular} & 19.04 & 17.18 & 16.51 & 13.10 & 15.88 & -\\ 
 \noalign{\smallskip}

 & Both & 26.64 & 4.30 & 22.68 & 25.83 & 23.14 &- \\ 
 \noalign{\smallskip}
\bottomrule[1pt]
 \noalign{\smallskip}

\end{tabular}

}
\end{center}
\end{table}
\setlength{\tabcolsep}{1.4pt}
The full quantitative experiment and more visualizations are included to demonstrate the effectiveness of the method.

\section{Quantitative Results}

Table~\ref{table:qr} shows the full results of the proposed UC-OWOD evaluation protocol. The detection performance of known classes is calculated by mAP. As mentioned above, \textit{Oracle} is a detector trained with annotations of all known and unknown instances. Since the training set only has labels of known classes in task 1, the detection result of \textit{Oracle} on unknown classes are not considered. Without finetuning, the model will completely forget previous classes, which results in significant mAP 
drop (55.38\% vs. 0\%). By finetuning, part of the detection ability of the preciously known classes can be restored (40.90\% mAP), but WI/A-OSE performance does suffer.
The finetuned detector is more inclined to classify an object into known classes. Regarding the scores about unknown classes, task 4 cannot be measured due to the lack of unknown ground-truth. On tasks with the previous known, our method
learns better than ORE on the current known. However, due to incomplete annotations of the validation set, detection of unknown objects such as \textit{house} are regarded as false detection. For this reason, Both mAP of our model is lower than the ORE. Therefore, mAP can only measure the detection performance of the model for known classes to a certain extent.

\begin{figure}[!t]
\centering
\includegraphics[width=0.95\textwidth]{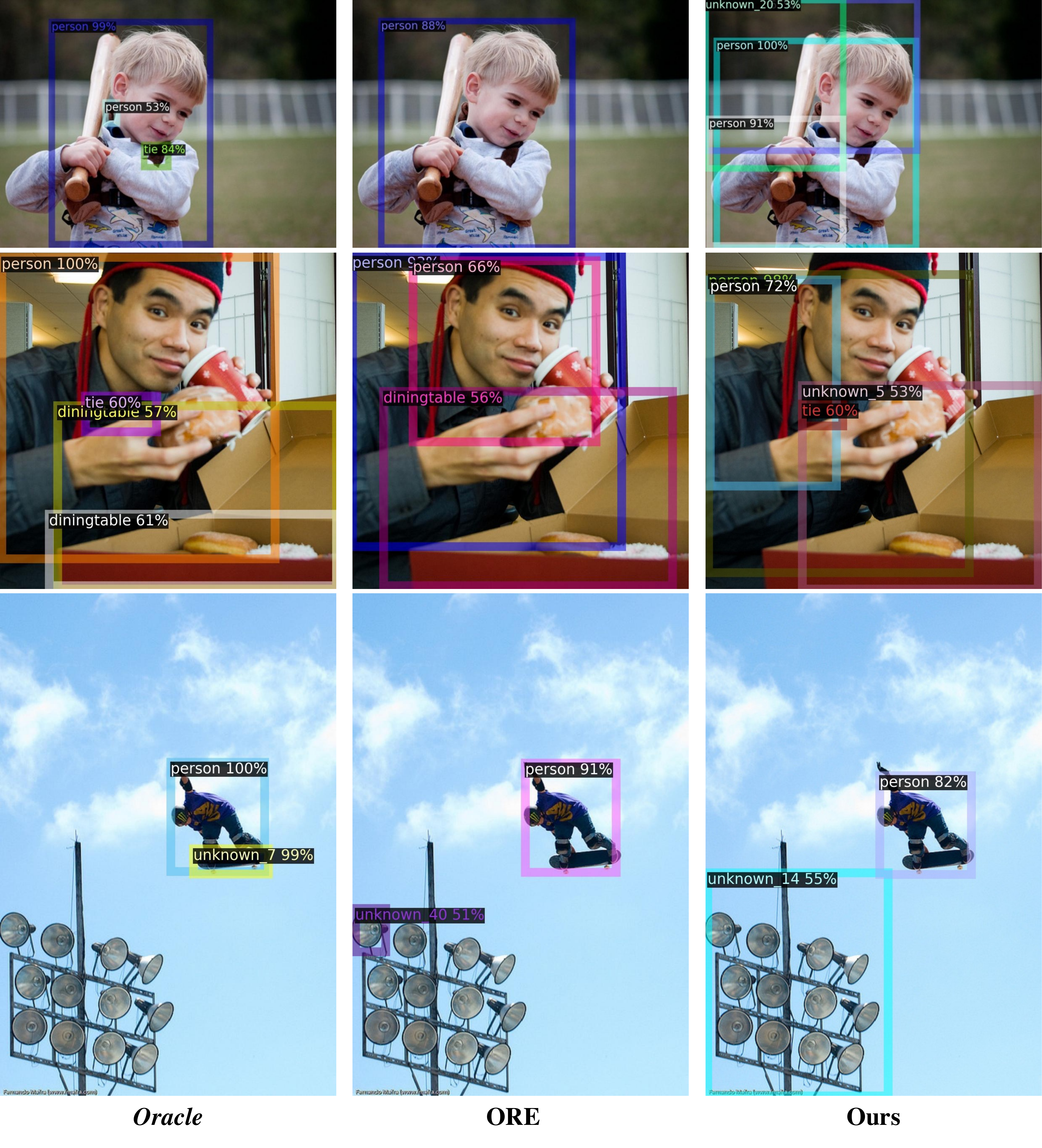}
\caption{
Detection results of \textit{Oracle}, ORE and our model. In the first row, \textit{Oracle} and ORE fail to detect the \textit{baseball~bat} in the image. In the second row, our model is able to correctly detect the \textit{donut}, while the other models mis-detect it as a \textit{dining table}. In the third row, our model and ORE can detect \textit{broadcast}, but the localization of our model is more accurate.}
\label{fig:com}
\end{figure}

\begin{figure}[!t]
\centering
\includegraphics[width=0.8\textwidth]{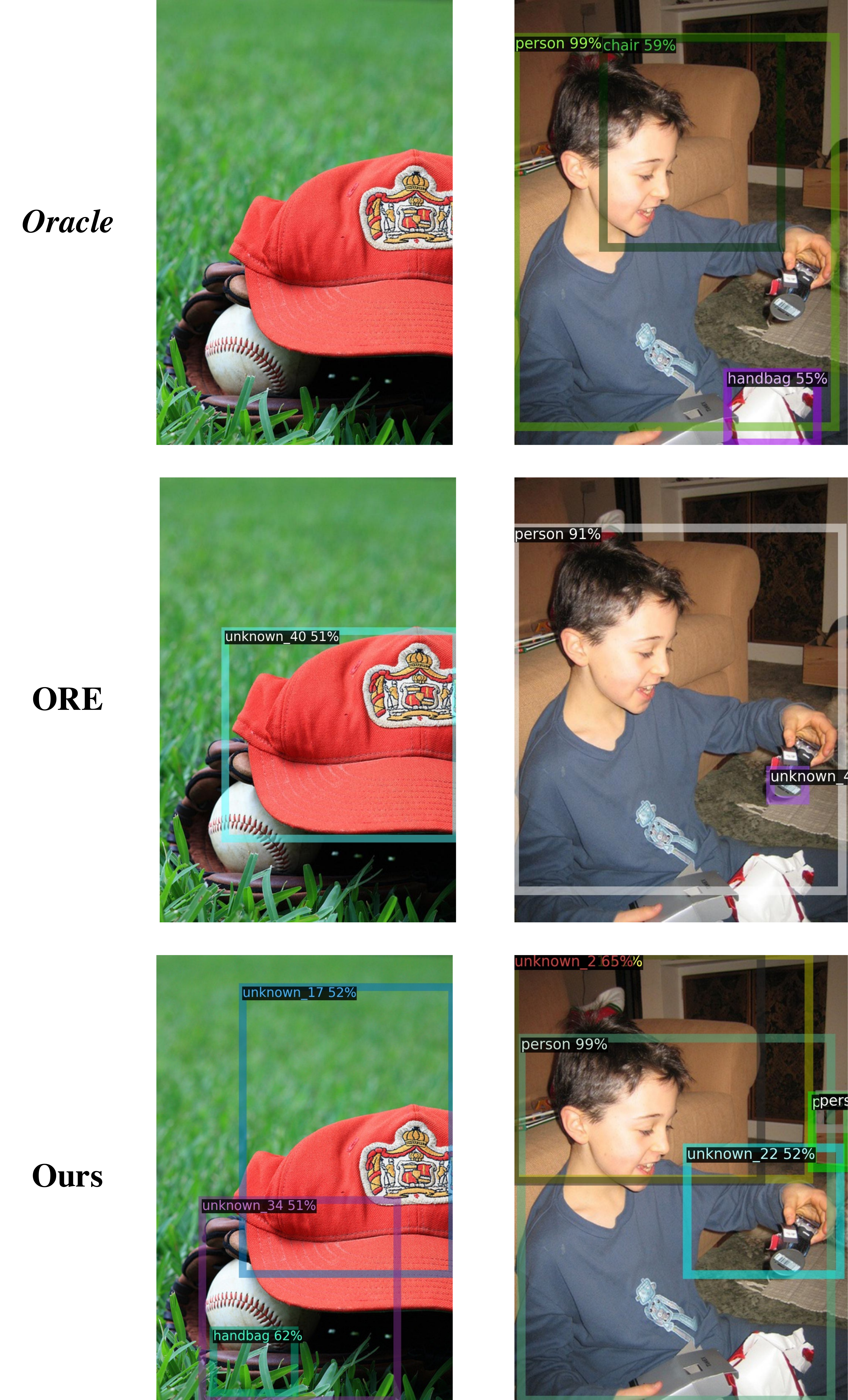}
\caption{
Detection results of multiple unknown objects. Only our model can correctly distinguish different unknown classes in an image.
}
\label{fig:ml}
\end{figure}

\begin{figure}[!t]
\centering
\includegraphics[width=1\textwidth]{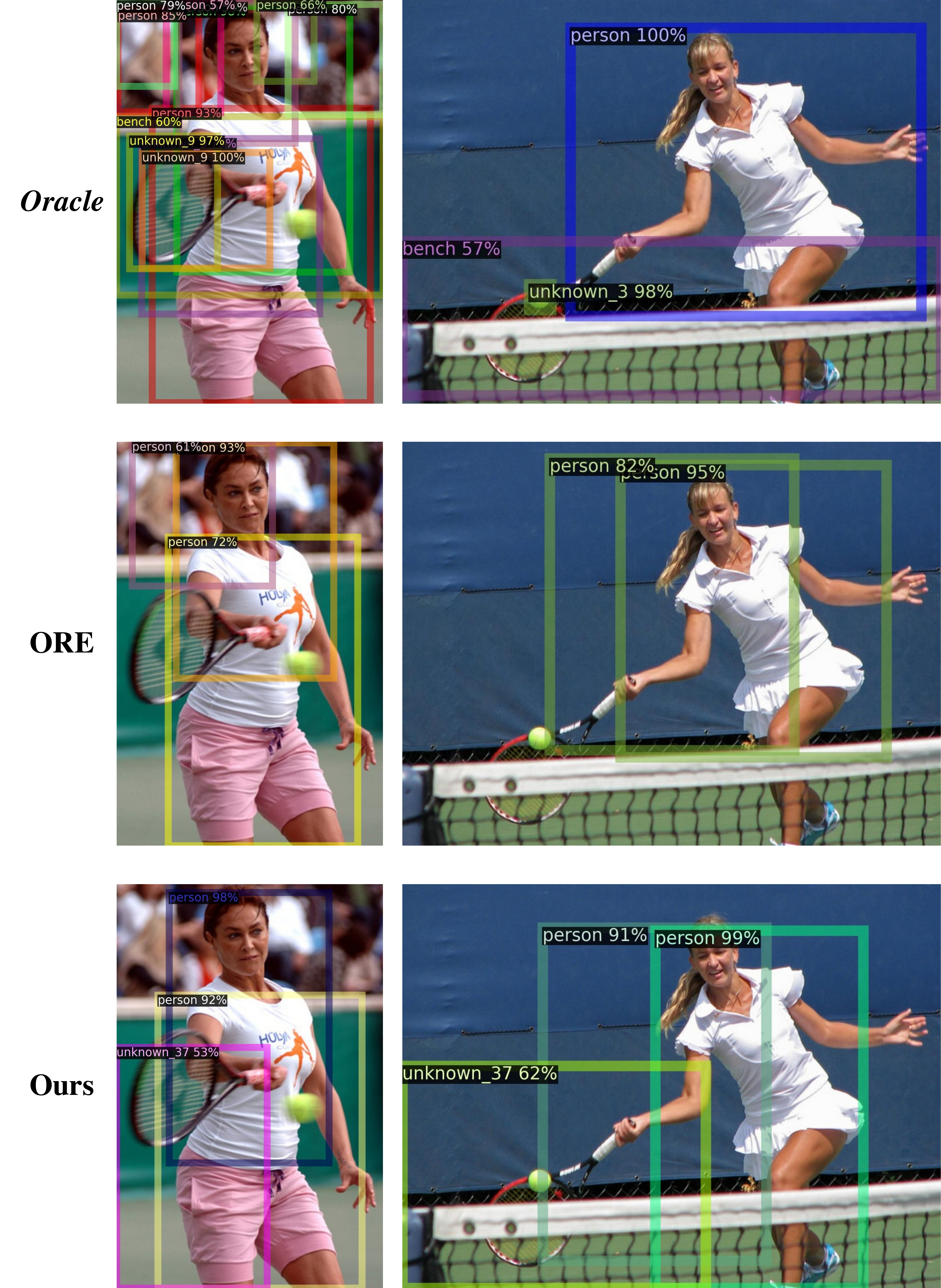}
\caption{
Detection results of unknown objects of the same class. Only our model can correctly locate unknown objects and classify them into the same unknown class.}
\label{fig:ten}
\end{figure}

\begin{figure}[!t]
\centering
\includegraphics[width=1\textwidth]{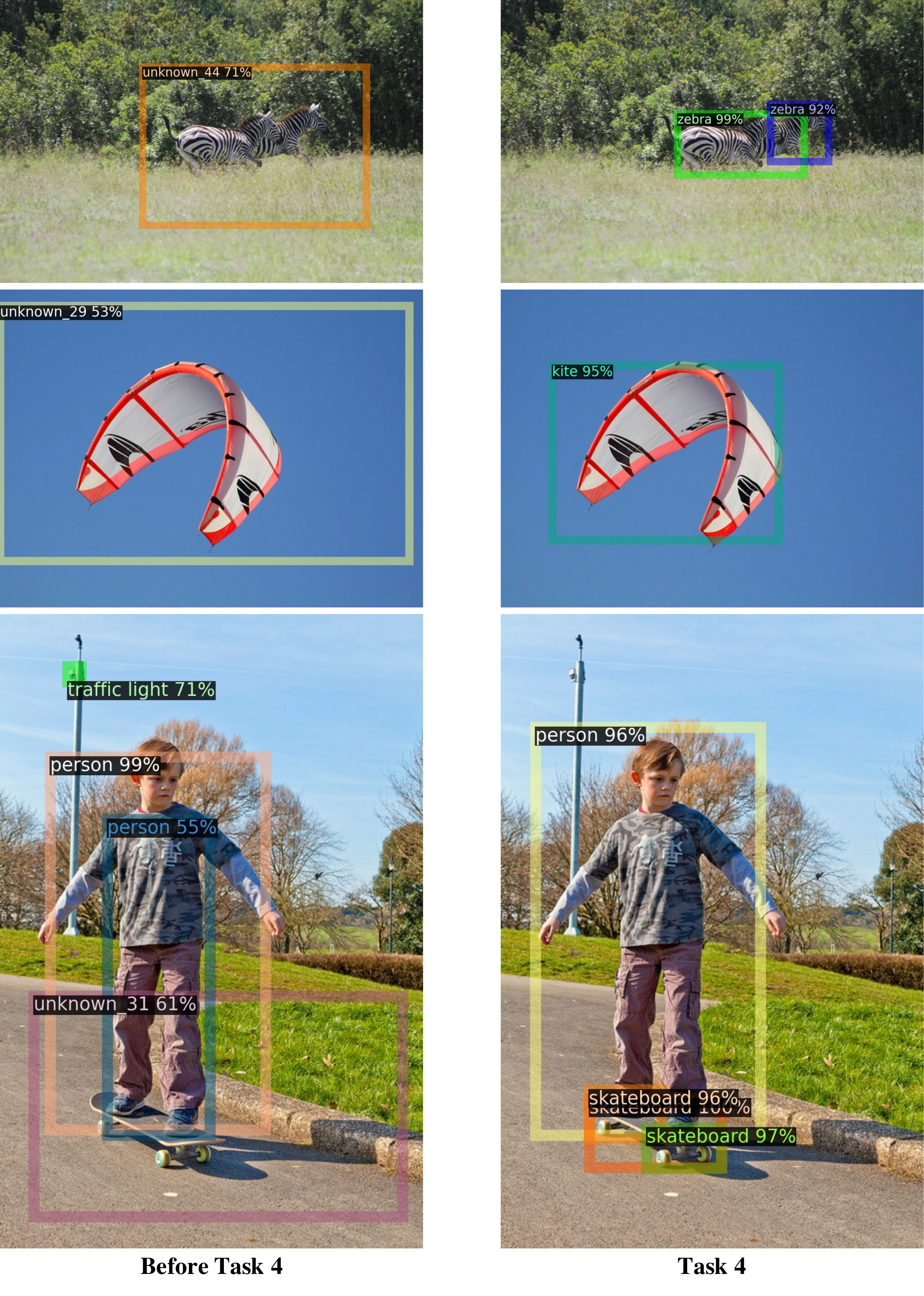}
\caption{The detection results of our model before task 4 are shown on the left. The corresponding predictions after incremental training using task 4 are shown on the right. In the first row, the \textit{unknown}-44 on the left is correctly predicted as \textit{zebra} in task 4. In the second row, the \textit{unknown}-29 is correctly detected as \textit{kite}. In the third row, task 4 correctly detects \textit{unknown}-31 as \textit{skateboard}.}
\label{fig:il}
\end{figure}
\section{Qualitative Results}
Since Faster-RCNN cannot detect any unknown objects, we only qualitatively compare \textit{Oracle}, ORE and our model, as shown in Fig.~\ref{fig:com}. For each test image, columns from left to right are the detection results of \textit{Oracle}, ORE, and our model. Both \textit{Oracle} and ORE failed to detect \textit{baseball~bat} and \textit{donut}, etc. This implies that our model is better at detecting unknown objects. 
In order to better analyze the performance of the model on the UC-OWOD problem, we use some images with multiple unknown instances to test, as shown in Fig.~\ref{fig:ml}. The results show that our model can correctly distinguish unknown objects, i.e., classifying \textit{baseball} as \textit{unknown}-34 and \textit{cap} as \textit{unknown}-17. In contrast, \textit{Oracle} and ORE can only detect unknown objects as one class. Fig.~\ref{fig:ten} shows the detection results of the same-class unknown objects on different images. Our model is able to detect 
\textit{tennis~racket} as \textit{unknown}-37 on different images, which both \textit{Oracle} and ORE fail to do. Fig.~\ref{fig:il} shows the qualitative results of incremental learning of our model on different tasks.
Our model is able to detect unknown objects and classify them as known classes when their labels are introduced, such as \textit{zebra}. 

\clearpage
%
%
\end{document}